\documentclass[conference]{IEEEtran}
\usepackage{amssymb}
\usepackage{amsmath}
\usepackage{amsthm}
\usepackage{graphicx}
\usepackage{subfigure}

\def\abstract{
\typeout{Abstract}
 {\bf Abstract} 
} 

\newcommand{\be}{\begin{equation}}
\newcommand{\ee}{\end{equation}}
\newcommand{\pai}{p_i^*}

\newcommand{\sset}[1]{\left\{#1\right\}}
\newcommand{\prob}[1]{\hbox{Pr}\left\{#1\right\}}
\newcommand{\he}{\hat{\epsilon}}
\newcommand{\eps}{\epsilon}
\newcommand{\fj}{\lfloor j+1 \rfloor}

\begin{document}

\title{Ensemble Validation:\\ Selectivity has a Price, but Variety is Free}

\author{Eric Bax \\ Verizon \\ baxhome@yahoo.com \and Farshad Kooti \\ Facebook}


\maketitle

\begin{abstract}
For an ensemble classifier that is composed of classifiers selected from a hypothesis set of classifiers, and that selects one of its constituent classifiers at random to use for each classification, we present ensemble error bounds consisting of the average of error bounds for the individual classifiers in the ensemble, a term that depends on the fraction of hypothesis classifiers selected for the ensemble, and a small constant term and multiplier. There is no penalty for using a richer hypothesis set, if the same fraction of the hypothesis classifiers are selected for the ensemble. 
\end{abstract}



\section{Introduction}
In machine learning, ensemble methods combine multiple classifiers into a single classifier. Ensemble methods include bagging \cite{breiman96}, boosting \cite{schapire90,freund97}, forests (of decision trees) \cite{ho95,ho98,breiman01}, and stacking \cite{wolpert92}. In this paper, we focus on the Gibbs ensemble classifier with a uniform distribution over its member classifiers. To classify an example, a Gibbs classifier chooses one of the classifiers from its ensemble at random and applies that classifier to the example. Our goal is to develop out-of-sample error bounds for Gibbs classifiers. 

We show that an ensemble with $s$ classifiers selected from $m$ hypothesis classifiers has essentially the same error bound (up to small constants) as a single classifier selected from a hypothesis set of $m$ classifiers. So, for example, essentially the same error bound applies to selecting 100 classifiers for an ensemble from a set of 10,000 classifiers as to selecting a single classifier from a set of 100, even though there are ${{10\,000}\choose{100}}$ ways to choose in the first case and only 100 in the second. In fact, the same bounds apply to selecting 1\% of the classifiers for an ensemble from an arbitrarily large hypothesis class. 

In contrast to early results on average error rates over classifiers \cite{bax98}, the bounds in this paper allow validation of ensemble classifiers based on data used to train other ensemble classifiers. For example, one method to develop an ensemble classifier is to split all in-sample data into training and validation data multiple times. Next, for each train-validation split, develop a classifier based on the training data and use the validation data to derive an error bound for the classifier -- yielding a hypothesis set of classifiers, each with a single-classifier error bound. Then select the classifiers with the lowest single-classifier error bounds to form an ensemble classifier that, for each example to be classified, selects one of its classifiers at random and applies it. More splits yields a richer hypothesis set. We show that there is no penalty for using a richer hypothesis set, if the same fraction of the hypothesis classifiers are selected for the ensemble.

Section \ref{related} surveys some related work. Section \ref{uniform} briefly reviews uniform and nearly uniform error bounds. In Section \ref{valens}, we show how to infer ensemble error bounds from uniform and nearly uniform error bounds. Section \ref{telescoping} shows how to expand nearly uniform error bounds into telescoping error bounds. Section \ref{selvar} shows how to select parameters for telescoping error bounds to make the price of variety independent of the hypothesis set size and the number of in-sample examples. In Section \ref{comps}, we compare ensemble error bounds. Section \ref{discussion} concludes with a discussion of how to apply the error bounds we derive. 

\section{Related Work} \label{related}
The PAC-Bayes technique \cite{mcallester99,langford01,begin16} bounds the average out-of-sample error rate over a set of classifiers. It applies to the following setup: start with a prior distribution over a set of hypothesis classifiers, use a set of in-sample data to select a posterior distribution over classifiers, and use that posterior as the basis for an ensemble classifier that selects one of its classifiers at random, according to the posterior, for each classification. PAC-Bayes error bounds include a term for selectivity that grows as the Kullback-Leibler (KL) divergence \cite{kullback51} (or another divergence \cite{begin16}) between the posterior and prior grows and a term that grows with the number of in-sample examples. For equally-weighted Gibbs ensemble classifiers and a uniform prior, the bounds in this paper remove the term that grows with sample size.

There is also prior work on validation of ensembles that combine outputs from their constituents, rather than selecting one at random. This includes validation of voting committees \cite{bax_voting} -- ensembles that classify by returning the most common output among their classifiers. For ensembles with continuous outputs, for function-fitting rather than classification, combining outputs is sometimes called fusion. There is a method to validate fusion using linear programming, with single-classifier error bounds giving constraints for the ensemble error bound \cite{bax_fusion}. There is also a method to select a convex combination of the classifier outputs for fusion, based on minimizing the ensemble error bound inferred from single-classifier error bounds for the classifiers in the ensemble \cite{bax_val_by_inference}.

\section{Uniform and Nearly Uniform Validation} \label{uniform}
Let's begin with a brief review of uniform validation. Let $m$ be the number of classifiers in the hypothesis class. Let $n$ be the number of validation examples, with known inputs and labels, available for each classifier in the hypothesis class. Assume the validation examples for each classifier are drawn independently of any data used to choose the classifier for the class and are drawn i.i.d. from an unknown input-label distribution $D$. (The validation data may be the same, partially overlap, or not overlap at all for different classifiers in the class.) For each classifier $i \in \sset{1, \ldots, m}$ in the hypothesis class, let $p_i$ be the error rate of classifier $i$ over its validation data. Let $\pai$ be the (unknown) error rate of classifier $i$ over $D$. Call $\pai$ the actual error rate or out-of-sample error rate of classifier $i$.

Applying Hoeffding bounds \cite{hoeffding63}, for each classifier $i$, for $\delta > 0$:
$$ \prob{\pai \geq p_i + \sqrt{\frac{\ln \frac{1}{\delta}}{2n}}} \leq \delta.$$
Substitute $\frac{\delta}{m}$ for $\delta$:
$$  \prob{\pai \geq p_i + \sqrt{\frac{\ln \frac{m}{\delta}}{2n}}} \leq \frac{\delta}{m}.$$
Use the sum bound for the probability of a union to derive a uniform bound:
$$ \prob{\exists i: \pai \geq p_i + \sqrt{\frac{\ln \frac{m}{\delta}}{2n}}} $$
\be \leq \sum_{i=1}^{m} \prob{\pai \geq p_i + \sqrt{\frac{\ln \frac{m}{\delta}}{2n}}} \leq m \frac{\delta}{m} = \delta.  \label{smile} \ee
This is the standard PAC error bound for uniform convergence of classifier validation error rates to their actual error rates, from Vapnik and Chervonenkis (VC) \cite{vapnik71}.

Uniform validation requires that, with probability at least $1 - \delta$, every classifier in the class has a successful validation in the sense that its validation error rate is close to its actual error rate. In contrast, nearly uniform validation \cite{bax08} allows some classifiers to have validation error rates far from their actual error rates. Call these misvalidations. 

\begin{figure} 
\includegraphics[width=3.5in]{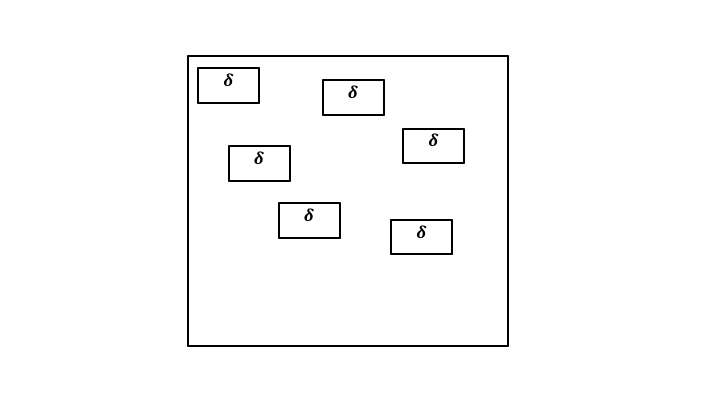} 
\caption{\textbf{Uniform Validation}. If each of 6 classifiers' error bounds have failure probability $\delta$, then the worst-case probability of at least one failure is $6 \delta$. The worst case is that failures are disjoint.}
\end{figure}

\begin{figure} 
\includegraphics[width=3.5in]{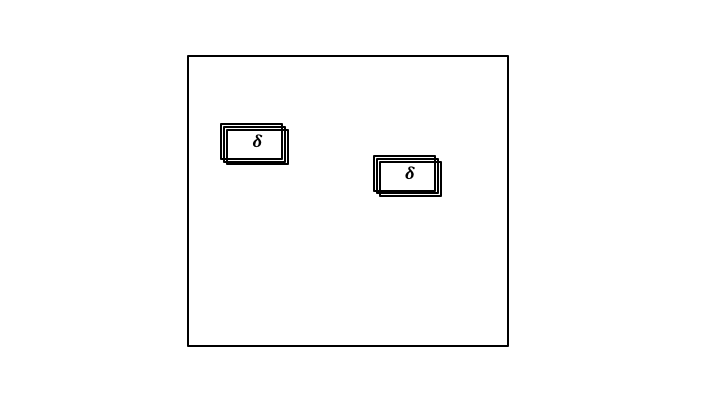} 
\caption{\textbf{Nearly Uniform Validation}. If each of 6 classifiers' error bounds have failure probability $\delta$, then the worst-case probability of at least three failures is $2 \delta$. The worst case is that any failure is simultaneous with two others.}\end{figure}

To derive nearly uniform error bounds, select a number $j$ of misvalidations to allow. The probability of more than $j$ misvalidations is maximized if there are exactly $j+1$ misvalidations in each case of any misvalidations. For $j=0$, this is just the sum bound for the union of probabilities: the probability of at least one misvalidation is maximized if there is exactly one misvalidation when there is any misvalidation -- any more and the probability of misvalidation will be less than the sum of individual classifier misvalidation probabilities. For $j=1$, which allows a single misvalidation, the greatest probability of two or more misvalidations is achieved if any misvalidation that occurs is accompanied by exactly one more.

So the probability of more than $j$ misvalidations is at most $\frac{1}{j+1}$ of the probability of one or more misvalidations.  Apply this rule to produce a nearly uniform version of Inequality \ref{smile}:
\be \prob{\left|\sset{i : \pai \geq p_i + \sqrt{\frac{\ln \frac{m}{\delta}}{2n}}}\right|>j} \leq \frac{\delta}{j+1}. \label{ok} \ee

We will use some non-integer values for $j$. If we allow a non-integer number $j$ of misvalidations, then the worst case is for $\fj$ misvalidations to occur simultaneously if there are more than zero misvalidations. So we will use $\fj$ in place of $j+1$ on the RHS of Inequality \ref{ok} to accommodate non-integer $j$ values. 

Substitute $\delta \fj$ for $\delta$:
\be  \prob{\left|\sset{i : \pai \geq p_i + \sqrt{\frac{\ln \frac{m}{\delta \fj}}{2n}}}\right|>j} \leq \delta. \label{nub}\ee
This is a nearly uniform PAC error bound. Comparing it to the uniform error bound in Inequality \ref{smile}, in exchange for allowing $j$ misvalidations, we can subtract $\ln \fj$ from the numerator in the square root.

\section{Validation for an Ensemble} \label{valens}
Let $S$ be a subset of the $m$ hypothesis classifiers. Refer to $S$ as the selected classifiers or the ensemble. Classifiers may be selected for the ensemble based on their validation error rates. Let $s = |S|$.

Our goal is to derive an error bound for the average out-of-sample error rate over the ensemble classifiers. Let $E_S$ be the expectation over ensemble classifiers. Specifically, let $E_S p_i$ be the average validation error rate, and let $E_S \pai$ be the average out-of-sample error rate. (Then $E_s \pai$ is the out-of-sample error rate of a Gibbs classifier based on a uniform distribution over the ensemble classifiers.)

For an average bound based on uniform bounds, multiply both sides of the inequality in the probability from Inequality \ref{smile} by $\frac{1}{s}$:
$$\prob{\exists i: \frac{\pai}{s} \geq \frac{1}{s} \left(p_i + \sqrt{\frac{\ln \frac{m}{\delta}}{2n}}\right)} \leq \delta.$$
So
$$\prob{\sum_{i \in S} \frac{\pai}{s} \geq \sum_{i \in S} \frac{1}{s} \left(p_i + \sqrt{\frac{\ln \frac{m}{\delta}}{2n}}\right)} \leq \delta,$$
and
\be \prob{E_S \pai \geq E_S p_i + \sqrt{\frac{\ln \frac{m}{\delta}}{2n}}} \leq \delta. \label{base} \ee

Now consider an average bound based on nearly uniform bounds. In the worst case (with the nearly uniform bound holding), the $j$ allowed misvalidations occur; they are all for classifiers in the ensemble, and those classifiers have the least validation error rates in the ensemble and out-of-sample error rates of one. In that case, each misvalidation adds $\frac{1}{s}$ to the average error bound, minus the contribution it would have made to the bound if not for misvalidation:
$$\frac{1}{s} \left(p_i + \sqrt{\frac{\ln \frac{m}{\delta}}{2n}}\right).$$
Let $L$ index the $j$ classifiers in $S$ with the least validation error rates. Then we have:
$$\prob{\sum_{i \in S} \frac{\pai}{s} \geq \sum_{i \in S} \frac{1}{s} \left(p_i + \sqrt{\frac{\ln \frac{m}{\delta \fj}}{2n}}\right) \right.$$
$$\left. + \frac{j}{s} - \sum_{i \in L} \frac{1}{s} \left(p_i + \sqrt{\frac{\ln \frac{m}{\delta \fj}}{2n}}\right)} \leq \delta.$$
Separate the terms in sums and collect the terms in square roots to get:
$$\prob{E_S \pai \geq E_S p_i + \left(1 - \frac{j}{s}\right) \sqrt{\frac{\ln \frac{m}{\delta \fj}}{2n}} \right. $$
$$\left. + \frac{j}{s} \left(1 - E_L p_i\right)} \leq \delta.$$
To produce a bound on the difference between average validation error rate and average out-of-sample error rate before observing the validation error rates, consider that in the worst case $\forall i \in L: p_i = 0$, so $E_L p_i = 0$:
\be \prob{E_S \pai - E_S p_i \geq \left(1 - \frac{j}{s}\right) \sqrt{\frac{\ln \frac{m}{\delta \fj}}{2n}} + \frac{j}{s}} \leq \delta. \label{bun} \ee
Let
$$\he(j, \delta) = \min\left(\sqrt{\frac{\ln \frac{m}{\delta \fj}}{2n}}, 1\right).$$
Then we can rewrite Inequality \ref{bun} as
\be \prob{E_S \pai - E_S p_i \geq \left(1 - \frac{j}{s}\right) \he(j, \delta) + \frac{j}{s}} \leq \delta. \label{ntb} \ee

\section{Telescoping Bounds} \label{telescoping}
In the bound in Inequality \ref{ntb}, we have assumed that misvalidated classifiers have error rate one, regardless of their validation error rates, because one is a trivial upper bound on out-of-sample error rate. Instead, we could use a loose bound for the misvalidations that is tighter than the trivial bound. Since this ``backstop" uniform bound must hold simultaneously with our original nearly uniform bound, we have to split $\delta$ between them and appeal to a sum bound on the probability of either of these bounds failing. Let $\delta_1 + \delta_2 = \delta$. Then we have:
$$\prob{E_S \pai - E_S p_i \geq \left(1 - \frac{j}{s}\right) \he(j, \delta_1) + \frac{j}{s} \he(0, \delta_2)} \leq \delta, \label{tele1}$$
where $\he(0, \delta_2)$ is the backstop uniform bound. 

Extending this idea produces ``telescoping" bounds: a uniform bound and a sequence of nearly uniform bounds, each one tighter but allowing more misvalidations than the previous one. Each bound's $\he$ applies to the misvalidations allowed by the next bound but not by the present one. The final bound applies to the classifiers that are not misvalidated. Listing the bounds from right to left (it is math, after all) gives the telescoping bound:
$$\prob{E_S \pai - E_S p_i \geq \epsilon_t}$$
$$ \leq \delta_1 + \ldots + \delta_{t+1}, $$
where
$$\epsilon_t = \left(1 - \frac{\sum_{i=1}^{t} j_i}{s}\right)\he(\sum_{i=1}^{t} j_i, \delta_1) + \frac{j_1}{s} \he(\sum_{i=2}^{t} j_i, \delta_2) $$
\be  + \ldots + \frac{j_{t-1}}{s} \he(j_t, \delta_t) + \frac{j_t}{s} \he(0,\delta_{t+1}).\label{tb}\ee
We can optimize this bound by selecting $j_1, \ldots, j_t$ and $\delta_1, \ldots, \delta_{t+1}$ through dynamic programming. For details, refer to Appendix \ref{dp}.

\section{The Price of Selectivity} \label{selvar}
Now we apply telescoping bounds to ensembles to prove that for all $c>0$ and $0<\delta\leq\frac{1}{e}$, there is an ensemble error bound with $\epsilon$ at most
$$\frac{1}{\sqrt{2n}} \left[ \sqrt{\ln \frac{m}{s} + \ln \frac{1}{\delta}} \left(\frac{e^c}{e^c - 1}\right) \right. $$
\be \left. + \sqrt{c + 1} \left(\frac{e^c}{e^c - 1}\right)^2  + 1\right]. \label{bigresult} \ee
For $c=3$, for example, this gives
\be \epsilon_* \leq \frac{1}{\sqrt{2n}} \left[ 1.06 \sqrt{ \ln \frac{m}{s} + \ln \frac{1}{\delta}} + 3.22\right]. \label{estar3} \ee

The term $\ln \frac{m}{s}$ can be viewed as a price for selectivity: for a single classifier ($s=1$), it gives the $\ln m$ term from the standard VC error bounds \cite{vapnik71} for selecting one classifier from a size-$m$ hypothesis class (our Inequality \ref{smile}). For an average over all classifiers in the hypothesis class ($s=m$), it gives $\ln \frac{m}{m} = \ln 1 = 0$, which is consistent with a VC error bound for a hypothesis class consisting of a single hypothesis  -- the Gibbs classifier composed of all hypothesis classifiers. (For more details refer to Bax \cite{bax98}.) 

To prove the result, begin with the telescoping bound range from Inequality \ref{tb}:
$$ \epsilon_t = \left(1 - \frac{\sum_{i=1}^{t} j_i}{s}\right)\he(\sum_{i=1}^{t} j_i, \delta_1) + \frac{j_1}{s} \he(\sum_{i=2}^{t} j_i, \delta_2) $$
$$ + \ldots + \frac{j_{t-1}}{s} \he(j_t, \delta_t) + \frac{j_t}{s} \he(0,\delta_{t+1}). $$
Note that
$$\he(j, \delta) = \min\left(\sqrt{\frac{\ln \frac{m}{\delta \fj}}{2n}}, 1\right) \leq  \min\left(\sqrt{\frac{\ln \frac{m}{\delta j}}{2n}}, 1\right).$$
Also, $\forall j' \leq j: \he(j',\delta) \leq \he(j,\delta)$, so
$$\forall 1 \leq h \leq t: \he(\sum_{i=h}^{t} j_i, \delta_h) \leq \he(j_h, \delta_h).$$
For convenience, define $j_0 = s$. Now use $i$ as the index of summation over terms in $\epsilon_t$ (instead of over $j$ values within each term). Then
\be \epsilon_t \leq \sum_{i=1}^{t} \frac{j_{i-1}}{s} \he(j_i, \delta_i) + \frac{j_t}{s} \he(0, \delta_{t+1}).\label{qq} \ee

Let $c>0$ be a constant we select. Let $t$ be the minimum integer such that 
$$ e^{ct} \geq \sqrt{2n},$$
in other words,
$$t = \lceil \frac{1}{2c} \ln 2n \rceil.$$
For $1 \leq i \leq t$, let $\delta_i = (e-1) \frac{\delta}{e^i}$. Let $\delta_{t+1} = 0$. For $1 \leq i \leq t$, let $j_i = \frac{s}{e^{ci}}$. Substitute these values into Inequality \ref{qq} to get a bound range, and call it $\epsilon_*$:
\be \epsilon_* = \sum_{i=1}^{t} \frac{1}{e^{c(i-1)}} \he(\frac{s}{e^{ci}}, (e-1) \frac{\delta}{e^i}) + \frac{1}{e^{ct}} \he(0,0). \label{estarr} \ee

By the definition of $t$, the last term is at most $\frac{1}{\sqrt{2n}}$. For each other term:
$$\frac{1}{e^{c(i-1)}} \he(\frac{s}{e^{ci}}, (e-1) \frac{\delta}{e^i})$$
$$ \leq \frac{1}{\sqrt{2n}} \sqrt{\ln \frac{m}{s \delta} + \ln e^{ci} + \ln (e-1) + \ln e^i} \left(\frac{1}{e^c}\right)^{i-1}$$
$$ \leq \frac{1}{\sqrt{2n}} \sqrt{\ln \frac{m}{s \delta} + \ln (e-1) + (ci + i)} \left(\frac{1}{e^c}\right)^{i-1}.$$
Since $\ln (e-1) < 1$, $\ln \frac{m}{s \delta} > 1$, and $ci + i > 1$, this is
$$ \leq \frac{1}{\sqrt{2n}} \sqrt{\ln \frac{m}{s \delta} + 2\sqrt{\left(\ln \frac{m}{s \delta}\right)(ci+i)} + (ci + i)} \left(\frac{1}{e^c}\right)^{i-1}$$
$$ = \frac{1}{\sqrt{2n}} \sqrt{\left(\sqrt{\ln \frac{m}{s \delta}} + \sqrt{ci + i}\right)^2} \left(\frac{1}{e^c}\right)^{i-1}$$
$$ = \frac{1}{\sqrt{2n}} \left[ \sqrt{\ln \frac{m}{s \delta}} \left(\frac{1}{e^c}\right)^{i-1} + \sqrt{ci + i} \left(\frac{1}{e^c}\right)^{i-1} \right]$$
$$ \leq \frac{1}{\sqrt{2n}} \left[ \sqrt{\ln \frac{m}{s \delta}} \left(\frac{1}{e^c}\right)^{i-1} + i \sqrt{c + 1} \left(\frac{1}{e^c}\right)^{i-1} \right].$$

Substituting these values into Equation \ref{estarr}, 
$$ \epsilon_* \leq \sum_{i=1}^{t} \frac{1}{\sqrt{2n}} \left[ \sqrt{\ln \frac{m}{s \delta}} \left(\frac{1}{e^c}\right)^{i-1} + i \sqrt{c + 1} \left(\frac{1}{e^c}\right)^{i-1} \right]$$
$$ + \frac{1}{\sqrt{2n}}.$$
$$ = \frac{1}{\sqrt{2n}} \left[ \sqrt{\ln \frac{m}{s \delta}} \sum_{i=1}^{t} \left(\frac{1}{e^c}\right)^{i-1} + \sqrt{c + 1} \sum_{i=1}^{t} i \left(\frac{1}{e^c}\right)^{i-1} \right. $$
\be \left.  + 1\right]. \label{estar_sums} \ee

For the first sum, apply the standard inequality:
$$\forall0 \leq x<1: \sum_{0}^{t-1} x^i \leq \frac{1}{1-x},$$
with $x = \frac{1}{e^c}$:
$$ \sum_{i=1}^{t} \left(\frac{1}{e^c}\right)^{i-1} \leq \frac{1}{1 - \frac{1}{e^c}} = \frac{e^c}{e^c - 1}.$$

The second sum is:
$$1 + 2x + 3x^2 + \ldots $$
$$ = (1 + x + x^2 + \ldots) + (x + x^2 + x^3 + \ldots) + (x^2 + x^3 + x^4 + \ldots) + \ldots$$
$$ = (1 + x + x^2 + \ldots) + x (1 + x + x^2 + \ldots) $$
$$ + x^2 (1 + x + x^2 + \ldots) + \ldots$$
$$ = (1 + x + x^2 + \ldots) (1 + x + x^2 + \ldots)$$
$$ \leq \left(\frac{1}{1-x}\right)^2.$$
So
$$ \sum_{i=1}^{t} i \left(\frac{1}{e^c}\right)^{i-1} \leq \left(\frac{1}{1 - \frac{1}{e^c}}\right)^2 = \left(\frac{e^c}{e^c - 1}\right)^2.$$

Substitute these values for the sums in Expression \ref{estar_sums} to show that
$$ \epsilon_* \leq \frac{1}{\sqrt{2n}} \left[ \sqrt{\ln \frac{m}{s} + \ln \frac{1}{\delta}} \left(\frac{e^c}{e^c - 1}\right) \right. $$
\be \left. + \sqrt{c + 1} \left(\frac{e^c}{e^c - 1}\right)^2  + 1\right]. \ee

\section{Comparing Bounds} \label{comps}
Figures 1 and 2 compare error bound ranges $\epsilon$ for:
\begin{itemize}
\item Equation \ref{bigresult}, with the optimal $c$ value from $\{0.1, \ldots, 10\}$. (The optimal $c$ values are all between 2 and 4.)
\item Equation \ref{ntb}, with the optimal $j \in \{0, \ldots, s\}$. 
\item Equation \ref{estarr}, also with the optimal $c$ value from $\{0.1, \ldots, 10\}$. 
\item The``ideal" error bound: the (non-ensemble) uniform error bound over a class with $m/s$ classifiers: 
$$\frac{1}{\sqrt{2n}} \sqrt{ \ln \frac{m}{s} + \ln \frac{1}{\delta}}.$$
\end{itemize}

For $c=3$, Equation \ref{bigresult} is the ``ideal" error bound, multiplied by approximately $1.06$, and with $\frac{3.22}{\sqrt{2n}}$ added. Equation \ref{ntb} is the non-telescoping ensemble error bound derived by allowing $j$ misvalidations. Equation \ref{estarr} is the specific telescoping error bound that we used to derive Equation \ref{bigresult}. Since we bounded Equation \ref{estarr} to produce Equation \ref{bigresult}, Equation \ref{estarr} always gives a smaller error bound range than Equation \ref{bigresult}. 

Figure 1 shows how the bounds vary with selectivity: $\frac{s}{m}$. As the fraction of hypothesis classifiers selected for the ensemble decreases, the bound ranges increase -- more selective ensembles produce looser bounds. Equation \ref{estarr} gives the bound that is closest to the ideal bound.

Figure 2 shows how the bounds vary with variety: $m$. Equation \ref{bigresult} is constant over $m$ if $\frac{s}{m}$ is held constant. The other two bounds approach constant asymptotes with respect to $m$. Equation \ref{estarr} gives the bound that is closest to the ideal bound for large hypothesis class sizes. 

We recommend using Equation \ref{estarr} in practice. It is not as simple as Equation \ref{bigresult}, but it gives bounds that are significantly tighter. It is less complex than optimizing over the parameters for telescoping bounds (see Appendix \ref{dp}), and it gives bounds that are almost as tight as the optimal telescoping bounds.   

\begin{figure} 
\includegraphics[width=3.5in]{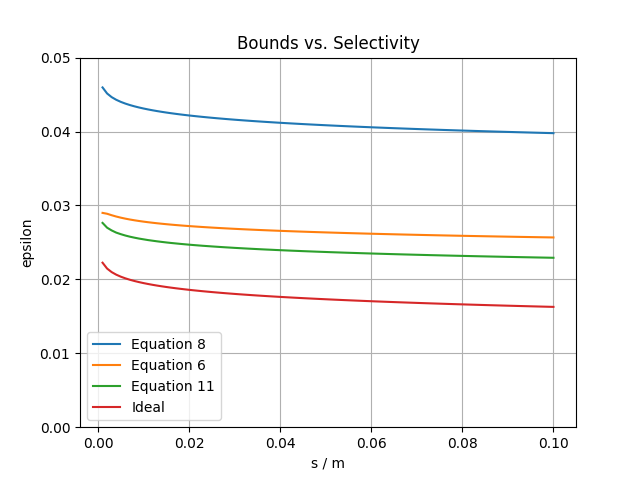} 
\caption{Equation \ref{estarr} is close to the ideal bound range over a wide range of selectivity: ensemble sizes as a fraction of hypothesis class sizes. Equation \ref{ntb} is similar to Equation \ref{estarr}. ($n = 10\,000$, $\delta = 0.05$, $m = 1\,000\,000$.)}
\end{figure}

\begin{figure} 
\includegraphics[width=3.5in]{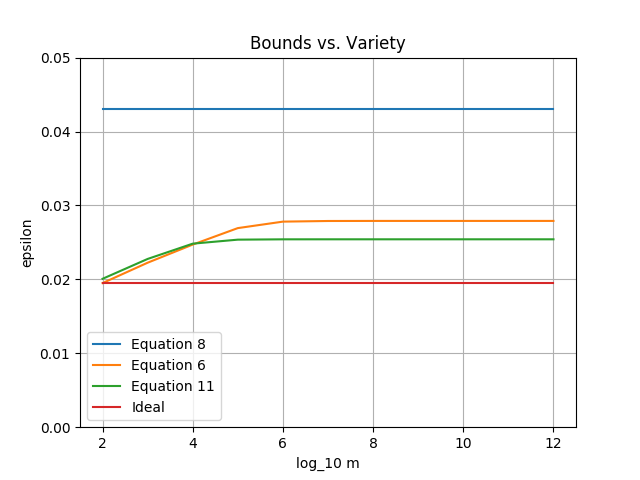}
\caption{Results for a large range of variety: hypothesis class sizes $m$, with selectivity held constant: $\frac{s}{m} = 0.01$. (Note that the horizontal axis is log-scale.) Equation \ref{estarr} is closest to the ideal bound range for large hypothesis class sizes $m$, and close to Equation \ref{ntb} for smaller hypothesis class sizes. Equation \ref{bigresult} does not vary with $m$ if $\frac{s}{m}$ is held constant. The other two equations vary only because of discretization effects: $\fj$ in $\hat{\epsilon}$ for Equation \ref{estarr} and also $j$ restricted to $\{0, \ldots, s\}$ for Equation \ref{ntb}. As $m$ increases, these effects diminish, and the bounds quickly approach asymptotes that are constant with respect to $m$. ($n = 10\,000$, $\delta = 0.05$.)}
\end{figure}

\section{Discussion} \label{discussion}
The error bounds developed in this paper allow the validation data for one classifier to be used to choose or train other classifiers for the hypothesis class. So we can apply these bounds to a Gibbs classifier based on classifiers trained on different splits of the same in-sample data set, with each classifier's validation data the data withheld from its training. Then we can extend an error bound for this holdout Gibbs classifier to an error bound for the single ``full" classifier based on all in-sample examples, if we can bound the rate of disagreement between the holdout Gibbs classifier and the full classifier, since the error rate of the full classifier is at most the error rate of the holdout Gibbs classifier plus the rate of disagreement. (For more details, see Bax and Le \cite{bax15}.) 

For local classifiers, like nearest neighbor classifiers, we can use statistics to bound the rate of disagreement. For example, suppose we split the in-sample data into 10 subsets; use each subset as the validation data for a one-nearest neighbor classifier consisting of the other nine subsets; use those ten classifiers in a Gibbs classifier, and use the results in this paper to compute an error bound for the Gibbs classifier. To get an error bound for the one-nearest neighbor classifier based on all in-sample examples, we also need to bound the rate of disagreement between that classifier and the Gibbs classifier. The rate of disagreement is at most 10\%, because with probability 90\%, the Gibbs classifier selects one of the nine classifiers that includes the nearest neighbor to the example being classified. (In the other 10\% of cases, the Gibbs classifier selects the one classifier with the nearest neighbor in its withheld validation data.)

For the Gibbs classifier based on only $m=10$ classifiers, we could use the average bound based on uniform validation, from Inequality \ref{base}. Alternatively, to reduce variance, we could use a Gibbs classifier based on all subsets of 90\% of the data. It would also have a 10\% bound on its rate of disagreement with the full classifier, because this Gibbs classifier also has a 10\% probability of selecting a classifier that does not include the nearest neighbor from the full classifier. For this Gibbs classifier, it is important to use error bounds that have a low cost for variety, because there are $m = {{10 n}\choose{n}}$ classifiers in the hypothesis set, where $10 n$ is the total number of in-sample examples and $n$ is the size of each withheld validation set. So we should use the bound from Equation \ref{estarr}. (The cost for selectivity is zero, since $s = m$.) For an efficient method to compute the average validation error rate for this Gibbs classifier, see \cite{bax_rknn}.

For non-local classifiers, where removal of examples can have non-local effects on classification, there are a few options to bound rates of disagreement between holdout Gibbs classifiers and full classifiers, depending on data availability. If the out-of-sample example inputs are available (the transductive setting \cite{vapnik98}), then compute the rate of disagreement directly over those examples. Otherwise, if there are unlabeled examples available from the out-of-sample input space, then the bound the rate of disagreement using standard error bounds, counting disagreements instead of errors. 

In the future, it would be interesting to determine whether nearly uniform bounds can improve PAC-Bayesian bounds that are based on arbitrary prior and posterior distributions and on change of measure with respect to divergences. A first step would be to show how nearly uniform bounds apply to arbitrary distributions. Then we would need to show that those improvements can be maintained through a change of measure.

Finally, the bounds in this paper rely on Hoeffding bounds \cite{hoeffding63}. There are alternatives to Hoeffding bounds that are tighter for accurate classifiers or that are sharp but require more computation. The type of ensemble bounds developed in this paper can also be built on those alternatives; for details refer to Appendix \ref{altb}.

\bibliographystyle{unsrt}
\bibliography{bax}

\begin{thebibliography}{10}

\bibitem{breiman96}
Leo Breiman.
\newblock Bagging predictors.
\newblock {\em Machine Learning}, 24(2):123--140, 1996.

\bibitem{schapire90}
Robert~E. Schapire.
\newblock The strength of weak learnability.
\newblock {\em Machine Learning}, 5(2):197--227, 1990.

\bibitem{freund97}
Yoav Freund and Robert Schapire.
\newblock A decision-theoretic generalization of on-line learning and an
  application to boosting.
\newblock {\em System Sciences}, 55(1):119--139, 1997.

\bibitem{ho95}
Tin~Kam Ho.
\newblock Random decision forests.
\newblock {\em Proceedings of the 3rd International Conference on Document
  Analysis and Recognition, Montreal, QC}, pages 278--282, 1995.

\bibitem{ho98}
Tin~Kam Ho.
\newblock The random subspace method for constructing decision forests.
\newblock {\em EEE Transactions on Pattern Analysis and Machine Intelligence},
  20(8):832--844, 1998.

\bibitem{breiman01}
Leo Breiman.
\newblock Random forests.
\newblock {\em Machine Learning}, 45(1):5--32, 2001.

\bibitem{wolpert92}
D.~Wolpert.
\newblock Stacked generalization.
\newblock {\em Neural networks}, 5(2):241--259, 1992.

\bibitem{bax98}
E.~Bax.
\newblock Validation of average error rate over classifiers.
\newblock {\em Pattern Recognition Letters}, pages 127--132, 1998.

\bibitem{mcallester99}
David~A. McAllester.
\newblock Pac-bayesian model averaging.
\newblock In {\em In Proceedings of the Twelfth Annual Conference on
  Computational Learning Theory}, pages 164--170. ACM Press, 1999.

\bibitem{langford01}
John Langford, Matthias Seeger, and Nimrod Megiddo.
\newblock An improved predictive accuracy bound for averaging classifiers.
\newblock In {\em In Proceeding of the Eighteenth International Conference on
  Machine Learning}, pages 290--297, 2001.

\bibitem{begin16}
Luc Begin, Pascal Germain, Francois Laviolette, and Jean-Francis Roy.
\newblock Pac-bayesian bounds based on the renyi divergence.
\newblock {\em Proceedings of the 19th International Conference on Artificial
  Intelligence and Statistics (AISTATS)}, 2016.

\bibitem{kullback51}
S.~Kullback and R.~A. Leibler.
\newblock On information and sufficiency.
\newblock {\em Annals of Mathematical Statistics}, 22(1):79--86, 1951.

\bibitem{bax_voting}
Eric Bax.
\newblock Validation of voting committees.
\newblock {\em Neural Computation}, 10(4):975--986, 1998.

\bibitem{bax_fusion}
E.~{Bax}.
\newblock Validation of fusion through linear programming.
\newblock In {\em IJCNN'99. International Joint Conference on Neural Networks.
  Proceedings (Cat. No.99CH36339)}, volume~1, pages 572--575 vol.1, July 1999.

\bibitem{bax_val_by_inference}
E.~Bax.
\newblock Using validation by inference to select a hypothesis function.
\newblock In {\em Pattern Recognition, International Conference on}, volume~2,
  page 2700, Los Alamitos, CA, USA, sep 2000. IEEE Computer Society.

\bibitem{hoeffding63}
W.~Hoeffding.
\newblock Probability inequalities for sums of bounded random variables.
\newblock {\em Journal of the American Statistical Association},
  58(301):13--30, 1963.

\bibitem{vapnik71}
V.~Vapnik and A.~Chervonenkis.
\newblock On the uniform convergence of relative frequencies of events to their
  probabilities.
\newblock {\em Theory of Probability and its Applications}, 16:264--280, 1971.

\bibitem{bax08}
E.~Bax and A.~Callejas.
\newblock An error bound based on a worst likely assignment.
\newblock {\em Journal of Machine Learning Research}, 9:859--891, 2008.

\bibitem{bax15}
Eric Bax and Ya~Le.
\newblock Some theory for practical classifier validation.
\newblock {\em Baylearn}, 2015.

\bibitem{bax_rknn}
Eric Bax.
\newblock Rknn: Answer uncertainty with randomness for effective error bounds.
\newblock {\em submitted to IJCNN 2019}, 2018.

\bibitem{vapnik98}
V.~Vapnik.
\newblock {\em Statistical Learning Theory}.
\newblock John Wiley \& Sons, 1998.

\bibitem{audibert04}
J.-Y. Audibert.
\newblock {\em P{A}{C}-{B}ayesian Statistical Learning Theory}.
\newblock PhD thesis, Laboratoire de Probabilities et Modeles Aleatoires,
  Universites Paris 6 and Paris 7, 2004.

\bibitem{maurer09}
Andreas Maurer and Massimiliano Pontil.
\newblock Empirical bernstein bounds and sample-variance penalization.
\newblock {\em 22nd Annual Conference on Learning Theory (COLT)}, 2009.

\bibitem{bernstein37}
S.~N. Bernstein.
\newblock On certain modifications of {C}hebyshev's inequality.
\newblock {\em Doklady Akademii Nauk SSSR}, 17(6):275--277, 1937.

\bibitem{bennett62}
G.~Bennett.
\newblock Probability inequalities for the sum of independent random variables.
\newblock {\em Journal of the American Statistical Association},
  57(297):33--45, 1962.

\bibitem{hoel54}
P.~G. Hoel.
\newblock {\em Introduction to Mathematical Statistics}.
\newblock Wiley, 1954.

\bibitem{langford05}
J.~Langford.
\newblock Tutorial on practical prediction theory for classification.
\newblock {\em Journal of Machine Learning Research}, 6:273--306, 2005.

\bibitem{ross03}
T.~D. Ross.
\newblock Accurate confidence intervals for binomial proportion and {P}oisson
  rate estimation.
\newblock {\em Computers in Biology and Medicine}, 33:509--531, 2003.

\bibitem{agresti98}
A.~Agresti and B.~A. Coull.
\newblock Approximate is better than `exact' for interval estimation of
  biniomial proportions.
\newblock {\em The American Statistician}, 52:119--126, 1998.

\bibitem{fleiss03}
J.~L. Fleiss, B.~Levin, and M.~C. Paik.
\newblock {\em Statistical Methods for Rates and Proportions: Third Edition}.
\newblock Wiley, 2003.

\bibitem{boucheron13}
S.~Boucheron, G.~Lugosi, and P.~Massart.
\newblock {\em Concentration Inequalities -- A Nonasymptotic Theory of
  Independence}.
\newblock Oxford University Press, 2013.

\end{thebibliography}

\appendix
\subsection{How to Optimize Telescoping Bounds} \label{dp}
We can use dynamic programming to optimize (over a discrete set of candidate parameter values) the telescoping bound from Inequality \ref{tb}:
$$\prob{E_S \pai - E_S p_i \geq \left(1 - \frac{\sum_{i=1}^{t} j_i}{s}\right)\he(\sum_{i=1}^{t} j_i, \delta_1) \right. $$
$$\left. + \frac{j_1}{s} \he(\sum_{i=2}^{t} j_i, \delta_2) + \ldots + \frac{j_{t-1}}{s} \he(j_t, \delta_t) + \frac{j_t}{s} \he(0,\delta_{t+1})}$$
$$ \leq \delta_1 + \ldots + \delta_{t+1}.$$
Let
$$\eps\left(\left(j_1, \ldots, j_t\right), \left(\delta_1, \ldots, \delta_{t+1}\right) \right) = $$
$$ \left(1 - \frac{\sum_{i=1}^{t} j_i}{s}\right)\he(\sum_{i=1}^{t} j_i, \delta_1) + \frac{j_1}{s} \he(\sum_{i=2}^{t} j_i, \delta_2) $$
\be + \ldots + \frac{j_{t-1}}{s} \he(j_t, \delta_t) + \frac{j_t}{s} \he(0,\delta_{t+1}). \label{estar} \ee
To optimize the bound, we want to compute:
$$\eps^*(\delta) = \min_{\forall i: j_i \in \sset{0, \ldots, s} \hbox{, } \sum_{i=1}^{t} j_i \leq s \hbox{, } \sum_{i=1}^{t+1} = \delta} $$
$$\eps\left(\left(j_1, \ldots, j_t\right), \left(\delta_1, \ldots, \delta_{t+1}\right) \right).$$
We will begin with the terms on the right of the RHS of Equality \ref{estar} and work to the left. Let
$$v(i, \Sigma_j, \Sigma_\delta) = $$
$$\min_{j_i + \ldots + j_t = \Sigma_j \hbox{, } \delta_{i+1} + \ldots + \delta_{t+1} = \Sigma_\delta}$$
$$ \left[ \frac{j_i}{s} \he(j_{i+1} + \ldots + j_t, \delta_{i+1}) \right.$$
$$ \left.+ \frac{j_{i+1}}{s} \he(j_{i+2} + \ldots + j_t, \delta_{i+2}) + \ldots + \frac{j_t}{s} \he(0, \delta_t) \right].$$
The base cases are:
$$ v(t, \Sigma_j, \Sigma_\delta) = \frac{\Sigma_j}{s} \he(0, \Sigma_\delta).$$
The general recurrence is:
$$v(i, \Sigma_j, \Sigma_\delta) =$$
$$ \min_{j_i \in \sset{0, \ldots, \Sigma_j} \hbox{, } \delta_{i+1} \in [0, \Sigma_\delta]} \frac{j_i}{s} \he(\Sigma_j - j_i, \delta_{i+1})$$
$$ + v(i+1, \Sigma_j - j_i, \Sigma_\delta - \delta_{i+1}).$$
(Use a discrete set of candidate values for $\delta_{i+1}$ in $[0, \Sigma_\delta]$, for example increments of 0.0001.) The last step is:
$$\eps^*(\delta) = \min_{\Sigma_j \in \sset{0, \ldots, s} \hbox{, } \delta_1 \in [0, \delta]} (1 - \frac{\Sigma_j}{s}) \he(\Sigma_j, \delta_1)$$
$$ + v(1, \Sigma_j, \delta - \delta_1).$$

\subsection{Using Alternative Single-Classifier Error Bounds} \label{altb}
We began with Hoeffding bounds \cite{hoeffding63} for single-classifier error bounds, then extended them to bounds for multiple classifiers, and then to bounds for multiple classifiers that allow some bound failures. Hoeffding bounds are the basis of many results in machine learning, because they have a form that is convenient for analysis, and because they were used in early and fundamental machine learning work on Vapnik-Chervonenkis (VC) dimension \cite{vapnik71} as a basis for error bounds over hypothesis classes. We can use other single-classifier bounds in place of Hoeffding bounds. There are many such bounds, including empirical Bernstein bounds \cite{audibert04} and bounds by Maurer and Pontil \cite{maurer09}, that are stronger for more accurate classifiers, based on Bernstein's bound \cite{bernstein37} and Bennett's bound \cite{bennett62}, which are stronger for distributions with smaller variances. (Hoeffding \cite{hoeffding63} also derived bounds that are stronger for more accurate classifiers.) There are also bounds that are sharp, but not as easily analyzed and require some computation, based on binomial inversion \cite{hoel54,langford05}. And there are various bounds from statistics \cite{ross03,agresti98,fleiss03}. For more on these single-classifier bounds for differences between sample means and actual means, refer to \cite{langford05} for an empirical comparison and to \cite{boucheron13} for in-depth exploration of concentration inequalities.

To substitute another single-classifier error bound for Hoeffding bounds, for bounds with ranges that do not depend on validation error rates (such as using the binomial inversion with the widest range for any error rate), let $\epsilon(\delta)$ be a function such that
$$ \forall \delta > 0: \prob{\pai \geq p_i + \epsilon(\delta)} \leq \delta.$$
Substitute $\frac{\delta}{m}$ for $\delta$:
$$  \prob{\pai \geq p_i + \epsilon(\frac{m}{\delta})} \leq \frac{\delta}{m}.$$
Then, for a bound over a set of classifiers, use the sum bound for the probability of a union:
$$ \prob{\exists i: \pai \geq p_i + \epsilon(\frac{m}{\delta})} $$
$$ \leq \sum_{i=1}^{m} \prob{\pai \geq p_i + \epsilon(\frac{m}{\delta})} $$
$$ \leq m \frac{\delta}{m} = \delta. $$

For nearly uniform validation allowing $j$ misvalidations, recall that having $j+1$ misvalidations in any case of at least one validation maximizes the probability of more than $j$ misvalidations. So that probability is at most $\frac{1}{j+1}$ of the probability of at least one misvalidation:
$$ \prob{\left|\sset{i : \pai \geq p_i + \epsilon(\frac{m}{\delta})}\right|>j} \leq \frac{\delta}{j+1}. $$
Now substitute $\delta \fj$ for $\delta$:
$$  \prob{\left|\sset{i : \pai \geq p_i + \epsilon(\frac{m}{\delta \fj})}\right|>j} \leq \delta. $$
This nearly uniform PAC error bound is a generalization of Inequality \ref{nub} -- one that allows other bounds in place of Hoeffding bounds.  To use it in place of Inequality \ref{nub} in our results for ensemble bounds, replace 
$$\he(j, \delta) = \min\left(\sqrt{\frac{\ln \frac{m}{\delta \fj}}{2n}}, 1\right) $$
by
$$\he(j, \delta) = \min\left(\epsilon(\frac{m}{\delta \fj}), 1\right). $$

For bounds with ranges that depend on validation error rates, for each classifier $i$ in the hypothesis class, let $\epsilon_i(\delta)$ be a function such that
$$ \forall \delta > 0: \prob{\pai \geq p_i + \epsilon_i(\delta)} \leq \delta.$$
Then
$$  \prob{\pai \geq p_i + \epsilon_i(\frac{m}{\delta})} \leq \frac{\delta}{m},$$
so
$$ \prob{\exists i: \pai \geq p_i + \epsilon_i(\frac{m}{\delta})} $$
$$ \leq \sum_{i=1}^{m} \prob{\pai \geq p_i + \epsilon_i(\frac{m}{\delta})} $$
$$ \leq m \frac{\delta}{m} = \delta. $$
For nearly uniform error bounds:
$$  \prob{\left|\sset{i : \pai \geq p_i + \epsilon_i(\frac{m}{\delta \fj})}\right|>j} \leq \delta. $$
For ensemble error bounds, the logic to derive Inequality \ref{bun} leads to a similar expression:
$$ \prob{E_S \pai - E_{S-J} p_i \geq E_{S - J} \epsilon_i(\frac{m}{\delta \fj}) + \frac{j}{s}} \leq \delta, $$
where $J$ indexes the set of $j$ ensemble classifiers having the least validation error bounds: $p_i +  \epsilon_i(\frac{m}{\delta \fj}).$
For telescoping error bounds, in Expression \ref{tb}, replace each $\he(j, \delta)$ by
$$ \he_S(j, \delta) = \min(E_{S - \hat{J}}  \epsilon_i(\frac{m}{\delta \fj}), 1), $$
where $\hat{J}$ indexes the set of $j$ ensemble classifiers having the least bound ranges: $\epsilon_i(\frac{m}{\delta \fj}).$

\end{document}